%% file: main.tex
\documentclass[sigconf, nonacm]{acmart}

\usepackage{booktabs}
\usepackage{graphicx}
\usepackage{subcaption}

\AtBeginDocument{%
  }

\setcopyright{none}

\usepackage{etoolbox}
\makeatletter
\AtBeginDocument{%
  \apptocmd{\ps@firstpagestyle}{\fancyfoot[L]{\footnotesize Preprint.}}{}{}%
}
\makeatother

\begin{document}

\title{The Regression Tax: Decomposing Why Skills Help --- and Hurt --- LLM Agents}
\titlenote{Sentient skill-creator skill: \url{https://github.com/sentient-agi/meta-skill-creator}}

\author{Darshan Tank}
\email{darshan@sentient.xyz}
\affiliation{%
  \institution{Sentient Labs}
  \country{}
}

\author{Baran Nama}
\email{baran@sentient.xyz}
\affiliation{%
  \institution{Sentient Labs}
  \country{}
}

\renewcommand{\shortauthors}{Darshan Tank and Baran Nama}

\begin{abstract}
Adding procedural skills to an LLM agent is typically evaluated by average
improvement in task success. However, this metric hides an important cost:
skills can also make agents worse. We measure both sides by comparing agents
with and without skills across nearly 6,000 runs spanning two office-automation
benchmarks and three model--harness stacks. This allows us to distinguish two
outcomes. A regression is a task solved without skills but failed after skills
are added. A residual failure is a task that fails both with and without skills.
We find that regressions are substantial enough that the best-performing skills
outperform others primarily by regressing less, not by gaining more. We identify
three causes of regression: (i) skill-description osmosis, a skill
changes an agent's behavior simply by being present in context, even when it is
never invoked; (ii) grounding displacement, a skill's prescribed procedure
overrides how the agent interprets its inputs; and (iii) verification
displacement, where the procedure suppresses checks the agent would otherwise
perform on its outputs. Analysing persistent failures reveals the same
underlying pattern. Existing skills overemphasize procedural guidance---the
stage least often responsible for failure---while under-supporting grounding and
verification, the dominant sources of remaining errors. After correcting
evaluation artifacts and studying traces, we find many regressions and persistent
failures recoverable through better grounding and verification. Procedural skills
should be evaluated by decomposing their net effect into gains and regressions,
not by aggregate improvement alone. We identify
three regression modes skills should avoid, and find that reliability depends more on
grounding and verification than on procedural skill choice.
\end{abstract}

\maketitle

\input{sections/01-introduction}
\input{sections/02-related-work}
\input{sections/03-setup}
\input{sections/04-gains-regressions}
\input{sections/05-mechanisms}
\input{sections/06-discussion}
\input{sections/07-conclusion}

\bibliographystyle{ACM-Reference-Format}
\bibliography{references}

\appendix
\section*{Appendix}
\input{sections/A1-appendix}

\end{document}

%% file: sections/01-introduction.tex
\section{Introduction}
\label{sec:intro}

An agent skill is a reusable piece of procedural guidance: a short description, a body
of natural-language instructions, and sometimes bundled code. It is loaded into the
agent's context to control how the agent performs a task. Recent systems such as
Trace2Skill, EvoSkill, and SkillOpt generate and optimise skills automatically from agent execution
traces~\cite{trace2skill,evoskill,skillopt}. These systems are evaluated by average
improvement in task success.

Average improvement alone is incomplete. Skills can also cause regressions: tasks the
agent solved without skills that it fails after skills are added. Two libraries can
have the same average gain and still differ in how many tasks they break.

Recent methods detect harmful skills and remove or suppress them~\cite{assay,grasp,rsea}.
This identifies which skills to drop, but not why a skill caused a regression, so a
mostly useful skill may still be discarded. These methods also act only when a skill
is retrieved or invoked. Skill's description remains in the system prompt on every step and can
change behavior without being called; selection-based methods do not measure that
effect. We address both: why regressions happen, and how to measure presence-only
effects.

\begin{figure*}[t]
  \centering
  \includegraphics[width=\textwidth]{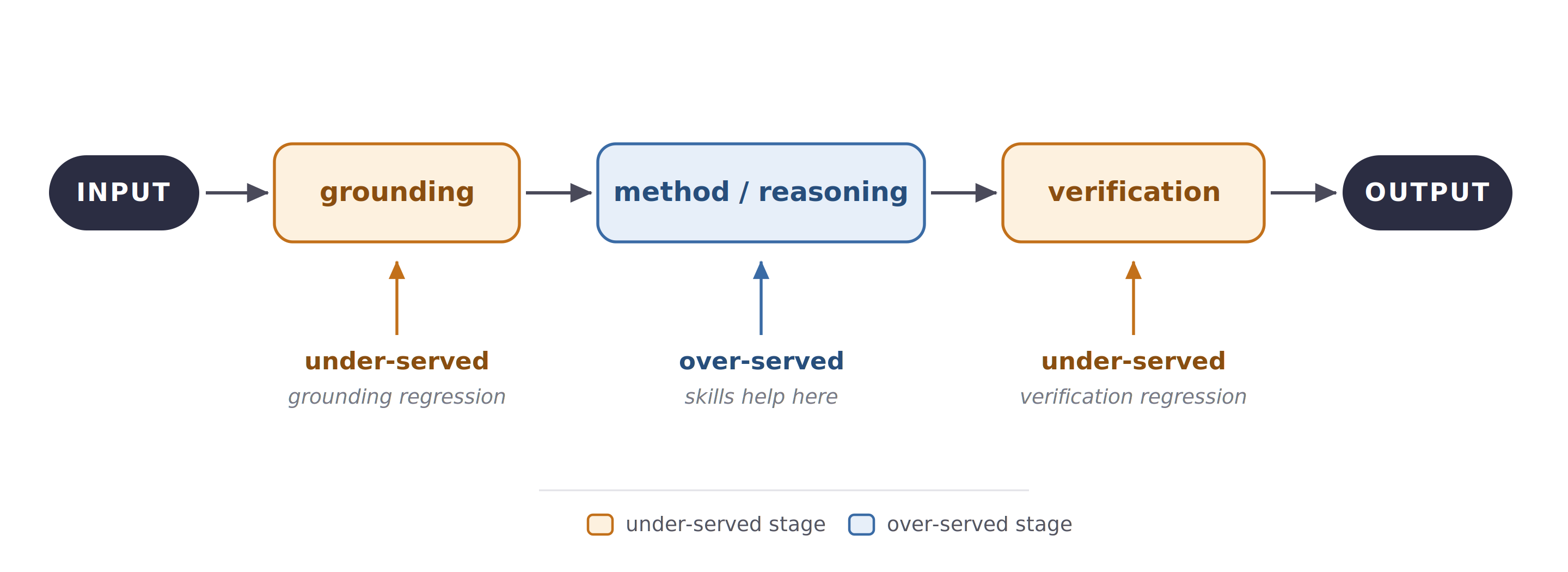}
  \caption{Three stages of an agent task: grounding (reading the right inputs),
    method (the procedure), and verification (checking the output). Existing skills
    mostly target the method stage. In our data, regressions and residual failures
    concentrate at grounding and verification instead.}
  \Description{A left-to-right pipeline diagram. INPUT feeds three stages in
    sequence---grounding, method or reasoning, and verification---and then OUTPUT.
    The grounding and verification stages are marked under-served and labelled as the
    sources of grounding and verification regressions; the middle method or reasoning
    stage is marked over-served and labelled where skills help. A legend distinguishes
    under-served from over-served stages.}
  \label{fig:axis}
\end{figure*}

We compare agents with and without skills across 5,832 task-condition runs on two
office-automation benchmarks and three model--harness stacks. Within each stack, only
the skill library changes. Because the pass-rate difference equals gains minus
regressions divided by the number of tasks, our contribution is not a new aggregate
metric but its paired decomposition: which failures a library fixes and which solved
tasks it breaks. The traces suggest three regression mechanisms---\emph{skill-description
osmosis}, \emph{grounding displacement}, and \emph{verification displacement}---and
place many residual failures at the grounding and verification stages
(Figure~\ref{fig:axis}).

\paragraph{Contributions.}
\begin{itemize}
  \item We decompose pass-rate changes into gains and regressions, showing that 324
    observed regression transitions offset 59\% of 553 gross gain transitions.
  \item We define and apply criteria for three candidate regression mechanisms,
    including description-only influence when no skill body is invoked.
  \item We locate persistent errors at grounding and verification, and re-grade 226
    treatment task-conditions that the original spreadsheet grader failed.
\end{itemize}

%% file: sections/02-related-work.tex
\section{Related Work}
\label{sec:related}

Several systems generate skills automatically from agent execution traces.
Trace2Skill turns trajectory lessons into transferable skills~\cite{trace2skill};
EvoSkill discovers skills for multi-agent systems~\cite{evoskill}; SkillOpt and
SkillOS optimize and maintain skill libraries~\cite{skillopt,skillos}. Meta-skill creators such as Anthropic's and OpenAI's skill-creator
tools~\cite{anthropic-skill-creator,openai-skill-creator} take the same idea further:
a skill that helps create new skills from failure signals.
These methods focus on creating skills and evaluate them by average gains. They do
not study the losses skills can cause.

\citet{huang2026datascience} evaluates LLM-generated skills on data-science tasks and
finds no reliable improvement over task-only prompting; gains and regressions occur in
roughly equal numbers across the tested conditions. The study shows that generic skill
guidance can conflict with task-specific instructions, but treats these cases as
diagnostic evidence rather than a mechanism account. We study this question in full
agent stacks and use execution traces to identify how regressions arise, including
effects from skill descriptions when no skill body is invoked.

Adding context is not free. \citet{shi2023distracted} show that irrelevant sentences
in a problem statement hurt reasoning on GSM-IC, and \citet{gsmdc} find that
controlled distractors corrupt both reasoning paths and arithmetic. Long-context work
shows accuracy drops when relevant information is buried in the middle of the
input~\cite{liu2023lost}, which surveys attribute to attention spreading thinner as
more tokens are added~\cite{memorysurvey}. A skill name and description in the system prompt are always
present, so it can change behavior even when never invoked. We later measure this
presence-only effect (osmosis).

Other work measures skill harm directly. ASSAY estimates each skill's effect with
randomized masking and masks skills predicted to hurt a task~\cite{assay}. GRASP
admits a skill only if it stays within a regression budget on a held-out
probe~\cite{grasp}. RSEA updates an evolved context layer only when held-out
performance does not drop~\cite{rsea}. SEAGym evaluates when self-evolution fails to
transfer~\cite{seagym}, and SkillGraph reduces redundant retrieval with a skill
dependency graph~\cite{skillgraph}. Across this line of work, regressions are handled
by selecting, masking, or dropping skills.

We take a different approach. Instead of only deciding which skills to keep, we ask
why a skill regresses a task, and we measure influence when the skill is never
retrieved---something selection methods do not cover.

%% file: sections/03-setup.tex
\section{Experimental Setup}
\label{sec:setup}

We run the same agent on the same tasks with and without skills and compare the
paired outcomes. This section covers the benchmarks, agent stacks, skill libraries,
outcome definitions, and significance test used in the rest of the paper. In each
comparison we hold everything fixed except the skill library in context.

\subsection{Benchmarks and Tasks}
\label{sec:setup-benchmarks}

We use two office-automation benchmarks over real artifacts: question answering
over financial documents, and spreadsheet manipulation. Both are utility tasks an
office worker would recognize, and both stress the parts of a task that skills are
meant to help with.

\textbf{OfficeQA-Pro} evaluates complex question answering over U.S.\ Treasury
financial documents. We curated and validated a challenging subset so measurement
stays reliable and evaluation focuses on hard cases. The benchmark asks an agent to
answer questions over long enterprise-style PDFs: retrieve facts from one or more
reports, read both text and tables, and do multi-step reasoning. Typical tasks
include extracting a figure or definition, comparing values across years or reports,
interpreting financial tables, computing sums, ratios, or percentage changes, and
combining information from several documents into one answer. The hard part is usually
grounding---finding the right table, vintage, and definition---not the arithmetic.
An answer is correct if it matches the ground-truth value within about one percent.

\textbf{SpreadsheetBench}~\cite{spreadsheetbench} evaluates spreadsheet manipulation
on real Excel-forum problems. Each task gives a workbook and a natural-language
instruction; the agent must edit the workbook correctly. Instructions cover locating
and filtering data, writing formulas, updating or filling cells, sorting and ranking,
removing duplicates, formatting, and working across multiple sheets. The benchmark
splits tasks into two types. \emph{Cell-level} tasks change a specified cell
range---for example extracting a substring into \texttt{B3:B14}, filling a column
with a formula, or updating a small set of target cells. \emph{Sheet-level} tasks
change a broader region or the sheet as a whole---for example building or rewriting a
table, applying an operation across several tables on one sheet, or coordinating edits
across sheets in a multi-sheet workbook. We keep tasks that can be graded by exact
cell value, and drop pure-formatting tasks plus a small set whose answer region uses
only uncached structured-reference formulas.

Cell-value grading has one known artifact. The grader compares target-cell values and
recalculates any formula the agent leaves in place. Its engine cannot evaluate Excel
structured or table references, or several modern functions, so a correct formula that
the engine cannot recompute is scored as a failure. This hurts agents that write
formulas rather than literal values. We leave the artifact in place for the main
results; in Section~\ref{sec:verification-displacement} we re-grade formula failures with a full
spreadsheet engine and report both raw and corrected scores.

\subsection{Model--Harness Stacks}
\label{sec:setup-stacks}

We run every benchmark on three model--harness stacks (Table~\ref{tab:stacks}). Each
stack pairs a harness with the model it normally uses. Harness and model are coupled,
so we treat the stack as the unit of analysis and discuss this as a limitation in
Section~\ref{sec:discussion}. Using three stacks lets us check whether effects hold
across setups rather than in only one.

\begin{table}[t]
  \begin{tabular}{ll}
    \toprule
    Harness      & Model \\
    \midrule
    OpenCode     & MiniMax-M2.7 \\
    Codex        & GPT-5.4-mini \\
    Claude Code  & Claude Sonnet 4.6 \\
    \bottomrule
  \end{tabular}
  \vspace{4pt}
  \caption{The three model--harness combinations we use, each run on both benchmarks
    under all four conditions.}
  \label{tab:stacks}
\end{table}

\subsection{Skill Libraries and How They Are Produced}
\label{sec:setup-pipeline}

Skill libraries are built with a four-stage pipeline, run separately for each stack.
(1) A \emph{baseline} run executes the agent with no skills and records trajectories.
(2) An \emph{analyst} reads those trajectories and extracts failure signals---recurring,
fixable causes of error. (3) A meta-skill creator turns the signals into skills,
with at most a few skills per signal. (4) An \emph{evaluation} run re-runs the
benchmark with those skills in context. The analyst runs once per stack, so all three
meta-skill creators---Anthropic's, OpenAI's, and Ours---start from the same signals.

The creators differ in how they turn those shared signals into skills, which is why
one signal set yields three different libraries.

The \textbf{anthropic} creator follows the Claude Code skill-authoring
guide~\cite{anthropic-skill-creator} and is measurement-driven. It drafts a skill,
runs the agent on test prompts with and without the skill, grades both runs, and
rewrites the skill from the measured difference, repeating until the skill helps. A
skill is kept only if a benchmarked run improved with it.

The \textbf{openai} creator follows the Codex guide~\cite{openai-skill-creator} and is
single-pass. It scaffolds the skill files, checks their structure with a validator,
and emits them. There is no eval loop and no measurement of a skill's effect.

The third creator, which we label \textbf{Ours}, is a harness-agnostic pipeline we
wrote for this study. Before writing, it searches the library for a similar skill and
updates that skill instead of adding a duplicate (discover-before-create). It then
drafts the skill and runs a self-critique pass that rewrites the draft against a
checklist. It uses only standard-library helpers and no external runs, subagents, or
CLI tools, so it can run in any harness.

The three creators therefore differ in what shapes the final text: a benchmarked eval
loop, a one-pass structural check, or self-critique with reuse. Creators are only a
way to get several libraries from one signal set. We do not rank them, and no claim
in this paper depends on which creator wrote a skill.

Each skill has a triggering description and a natural-language body
(Section~\ref{sec:intro}). The description stays in the system prompt on every step;
the body loads only when the skill is invoked. We use this split to separate effects
from invocation versus presence. Invocation is read from the trajectory.

Because the flows differ, the same signals expand or contract very differently. Within
a single stack the libraries range from 3 to 23 skills. On the
OpenCode\,$\cdot$\,minimax-m2.7 stack, for example, the openai creator turns the
signals into 23 skills, while Ours turns the same signals into 7. That a fixed
evidence set can take such different shapes is an early sign that how a skill is
written, not only what it encodes, affects outcomes.

\subsection{Conditions and the Controlled Comparison}
\label{sec:setup-conditions}

Each stack and benchmark is run under four conditions: \emph{none} (no skills), and
the three libraries \emph{anthropic}, \emph{openai}, and \emph{Ours}. Within a stack
and benchmark, only the library changes; the model, harness, task set, and analyst
signals stay fixed.

We use two comparisons. Treatment vs.\ \emph{none}, paired by task, measures the net
effect of adding a library (Section~\ref{sec:gains-regressions}). Treatment vs.\
treatment compares libraries built from the same signals and supports mechanism
attribution in Section~\ref{sec:mechanisms}.

Libraries are not length-matched: they differ in size and wording. A
treatment-versus-treatment contrast is therefore not a controlled token dose. We use
it only for individual tasks where we can trace the outcome to specific skill content.
We note this as a limitation in Section~\ref{sec:mechanisms}.

\subsection{Outcomes and Metrics}
\label{sec:setup-defs}

Each run yields a binary pass/fail from the benchmark grader. With paired with- and
without-skills runs, every task falls into one of four outcomes against the no-skill
baseline:

\begin{description}
  \item[Gain:] Failed in \emph{none}, passes with skills.
  \item[Regression:] Passed in \emph{none}, fails with skills.
  \item[Residual failure:] Failed in \emph{none}, still fails with skills.
  \item[Retained:] Passed in both.
\end{description}

The \emph{net effect} of a library is gains minus regressions against \emph{none}.
This is the main scalar we report; later sections also break regressions down by
mechanism.

\subsection{Scale and Statistical Protocol}
\label{sec:setup-protocol}

Each stack runs 486 tasks (94 OfficeQA-Pro + 392 SpreadsheetBench) under four
conditions, and there are three stacks, for
$486 \times 4 \times 3 = 5{,}832$ runs. These are paired comparisons on the same
tasks, not 5,832 independent tasks. Each condition is run once per task; we do not
replicate across seeds, so run-to-run variance is not estimated. We treat isolated
single-task flips carefully in Section~\ref{sec:mechanisms} and discuss variance in
Section~\ref{sec:discussion}.

Because treatments are paired with \emph{none}, we test net effect with the exact
McNemar test on discordant pairs (regressions and gains), two-sided. For each
condition we report regressions, gains, net, exact $p$-value, and a 95\% confidence
interval on the paired pass-rate difference. Tasks that pass in both or fail in both
drop out of the test. A library with many gains can still be non-significant if it
regresses nearly as many; that is why regression count matters as much as gain count
in Section~\ref{sec:gains-regressions}.

\begin{table*}[t]
  \footnotesize
  \setlength{\tabcolsep}{5pt}
  \begin{tabular}{llrrrrrl r}
    \toprule
    & & \multicolumn{2}{c}{Pass rate (\%)} & & & & & \\
    \cmidrule(lr){3-4}
    Stack & Library & without skill & with skill & Gains & Regressions & Net effect & $\Delta$ (95\% CI) & $p$ \\
    \midrule
    \multicolumn{9}{l}{\emph{OfficeQA-Pro} ($N=94$)} \\
    OpenCode\,$\cdot$\,minimax-m2.7
       & anthropic & 51.1 & 52.1 & 14 & 13 & $+1$          & $+1.1\ (-9.8,+11.9)$  & 1.000 \\
       & openai    & 51.1 & 55.3 & 16 & 12 & $+4$          & $+4.3\ (-6.7,+15.3)$  & 0.572 \\
       & Ours    & 51.1 & \textbf{60.6} & 15 &  6 & $+9$          & $+9.6\ (+0.2,+18.9)$  & 0.078 \\
    Codex\,$\cdot$\,gpt-5.4-mini
       & anthropic & 54.3 & 56.4 & 17 & 15 & $+2$          & $+2.1\ (-9.7,+13.9)$  & 0.860 \\
       & openai    & 54.3 & 56.4 & 11 &  9 & $+2$          & $+2.1\ (-7.2,+11.4)$  & 0.824 \\
       & Ours    & 54.3 & \textbf{57.4} & 16 & 13 & $+3$          & $+3.2\ (-8.0,+14.4)$  & 0.711 \\
    Claude Code\,$\cdot$\,sonnet-4.6
       & anthropic & 72.3 & \textbf{80.9} & 10 &  2 & $\mathbf{+8}$ & $+8.5\ (+1.5,+15.5)$  & 0.039 \\
       & openai    & 72.3 & 77.7 & 12 &  7 & $+5$          & $+5.3\ (-3.7,+14.3)$  & 0.359 \\
       & Ours    & 72.3 & 79.8 & 11 &  4 & $+7$          & $+7.4\ (-0.5,+15.4)$  & 0.118 \\
    \addlinespace
    \multicolumn{9}{l}{\emph{SpreadsheetBench} ($N=392$)} \\
    OpenCode\,$\cdot$\,minimax-m2.7
       & anthropic & 63.3 & \textbf{70.2} & 62 & 35 & $\mathbf{+27}$ & $+6.9\ (+2.0,+11.8)$ & 0.008 \\
       & openai    & 63.3 & 64.3 & 45 & 41 & $+4$           & $+1.0\ (-3.6,+5.7)$  & 0.747 \\
       & Ours    & 63.3 & 65.6 & 48 & 39 & $+9$           & $+2.3\ (-2.4,+7.0)$  & 0.391 \\
    Codex\,$\cdot$\,gpt-5.4-mini
       & anthropic & 65.8 & \textbf{67.6} & 30 & 23 & $+7$           & $+1.8\ (-1.9,+5.4)$  & 0.410 \\
       & openai    & 65.8 & 66.6 & 30 & 27 & $+3$           & $+0.8\ (-3.0,+4.5)$  & 0.791 \\
       & Ours    & 65.8 & 67.1 & 28 & 23 & $+5$           & $+1.3\ (-2.3,+4.8)$  & 0.576 \\
    Claude Code\,$\cdot$\,sonnet-4.6
       & anthropic & 70.2 & 81.1 & 59 & 16 & $\mathbf{+43}$ & $+11.0\ (+6.8,+15.2)$ & ${<}.001$ \\
       & openai    & 70.2 & 81.1 & 63 & 20 & $\mathbf{+43}$ & $+11.0\ (+6.5,+15.4)$ & ${<}.001$ \\
       & Ours    & 70.2 & \textbf{82.1} & 66 & 19 & $\mathbf{+47}$ & $+12.0\ (+7.5,+16.4)$ & ${<}.001$ \\
    \bottomrule
  \end{tabular}
  \vspace{6pt}
  \caption{Each library against the without-skill baseline, paired over the same
    tasks. The \emph{Pass rate} columns give the percentage of tasks passed without and
    with the library. \emph{Gains} ($c$) and \emph{Regressions} ($b$) are the counts
    defined in Section~\ref{sec:setup-defs}; \emph{Net effect} is $c-b$; $\Delta$ is
    the paired difference in pass rate in percentage points with a 95\% Newcombe
    confidence interval; $p$ is the exact two-sided McNemar test. A bold net effect
    marks significance at $p<.05$. Within each stack, and separately for each benchmark,
    the highest with-skill pass rate is also set in bold.}
  \label{tab:neteffect}
\end{table*}

%% file: sections/04-gains-regressions.tex
\section{Gains, Regressions, and Net Effect}
\label{sec:gains-regressions}

For a fixed task set, the pass-rate difference is exactly gains minus regressions,
divided by the number of tasks. The aggregate therefore already reports normalized net
effect; what it omits is the paired decomposition into tasks newly solved and tasks
newly broken. Table~\ref{tab:neteffect} reports both transition counts for all eighteen
library conditions.

\subsection{The Regression Tax}
\label{sec:tax}

Every library breaks some tasks the baseline had already solved. Across the eighteen
conditions, the regression count $b$ ranges from 2 to 41, and no condition has $b=0$.
Summed over all library conditions, we observe 553 gain transitions and 324 regression
transitions. Regressions therefore offset 59\% of gross gains, leaving a net of 229.
These are task-condition transitions rather than unique tasks: each library is paired
against the same no-skill condition. The pattern is similar on both benchmarks. On
OfficeQA-Pro the libraries gained 122 tasks and broke 81 (66\% of gains cancelled);
on SpreadsheetBench they gained 431 and broke 243 (56\% cancelled). This is not
limited to weak cells. Even when a library shows a clear positive delta, regressions
often sit close to gains: on SpreadsheetBench, OpenCode\,$\cdot$\,minimax-m2.7 openai
broke 41 while gaining 45, and Codex\,$\cdot$\,gpt-5.4-mini openai broke 27 while
gaining 30. Counting only the 553 gains, or only the pass-rate rise, misses much of
what the libraries did. That gap between gross gains and retained net improvement is
the \emph{regression tax} the paper is named for.

\subsection{Why the Decomposition Changes Interpretation}
\label{sec:regress-less}
\label{sec:metric}

The decomposition distinguishes libraries with similar aggregate performance. On
Claude Code\,$\cdot$\,sonnet-4.6 and OfficeQA-Pro, the three libraries gain 10, 11,
and 12 tasks but regress 2, 4, and 7. Ranking by gains puts openai first and anthropic
last; ranking by net effect reverses that order. Anthropic gains the fewest tasks but
has the largest net ($+8$) because it preserves more baseline successes. This pattern
is not universal: on SpreadsheetBench, the leading libraries often gain more as well
as regress less. The paired counts show which side drives each aggregate difference
without treating them as a separate metric.

\subsection{Statistical Uncertainty}
\label{sec:noise}

Five of the eighteen conditions reach nominal $p<.05$: Claude
Code\,$\cdot$\,sonnet-4.6 on SpreadsheetBench under all three libraries ($+43$, $+43$,
$+47$; all $p<.001$), the same stack's anthropic library on OfficeQA-Pro ($+8$,
$p=.039$), and OpenCode\,$\cdot$\,minimax-m2.7's anthropic library on SpreadsheetBench
($+27$, $p=.008$). The other thirteen confidence intervals include zero. Because these
are eighteen simultaneous comparisons, we apply a Bonferroni correction
($\alpha/18 = 0.0028$). Three conditions survive it---the Claude
Code\,$\cdot$\,sonnet-4.6 SpreadsheetBench trio, all with corrected $p<.001$---while
the two anthropic cells do not survive correction (adjusted $p=0.69$ and $p=0.14$).
The only net effect that holds up under correction is thus confined to a single stack and a
single benchmark: a real but narrow effect. This reinforces the regression-tax point. Were
skill libraries consistently converting added context into net gains, more of the eighteen
cells would remain significant after correcting for multiple comparisons; instead, most
apparent improvements are indistinguishable from noise once the correction is applied. We
therefore report the raw gain and regression counts as observed and treat them as a starting
point, reserving stronger causal interpretations for patterns supported by trajectory evidence
and cross-library contrasts.

%% file: sections/05-mechanisms.tex
\section{Mechanisms of Regression}
\label{sec:mechanisms}

We next use paired trajectories to classify how observed regressions arise. Each label
is grounded in paired-trajectory evidence and cross-library contrasts---the same task
flipping the same way across libraries---and reported as an observational classification
rather than a controlled causal test.

We do the full accounting on OfficeQA-Pro. Its answers are single numbers graded
against a ground truth with a numeric tolerance, so a regression is a real failure and 
the trajectory usually shows why. SpreadsheetBench regressions are less clean:
many are the value-versus-formula grader artifact from
Section~\ref{sec:setup-benchmarks}, and cell answers do not support the same
input-versus-output coding that separates grounding from verification. We therefore classify 
all 81 OfficeQA-Pro regressions with the full scheme (Table~\ref{tab:coverage}), and 
give SpreadsheetBench a coarser breakdown (Table~\ref{tab:sb-regressions}).

A regression is assigned by these criteria:
\begin{description}
  \item[Osmosis.] No skill body is invoked, yet the outcome changes, \emph{and}
        there is influence evidence: the same task flips the same way across
        libraries, or the behavioral change matches a concept named in a skill's
        description.
  \item[Grounding displacement.] A skill is invoked or read, and the error is at the
        input stage---wrong table, range, entity, definition, vintage, or
        premise---on a task the baseline had read correctly.
  \item[Verification displacement.] A skill is invoked or read, and the error is at
        the output stage---the procedure replaced or suppressed a check the baseline
        would have run on its own answer.
  \item[Other.] A presence-only flip with none of the influence evidence above, or a
        case without a readable trajectory.
\end{description}

For each regression, we compare the paired trajectories, record whether a skill body
was read, locate the first input- or output-stage divergence, and check whether the
same task changes similarly under another library. A single author assigned these
labels.

Table~\ref{tab:coverage} shows the result. Out of 81 regressions, 59 are grounding
displacement, 14 osmosis, 3 mixed grounding and verification, and 5 \emph{Other}.
Grounding dominates when a body is engaged; osmosis is concentrated when it is not.

\begin{table}[t]
  \footnotesize
  \setlength{\tabcolsep}{4pt}
  \begin{tabular}{lrrrrr}
    \toprule
    Mechanism & minimax-m2.7 & gpt-5.4-mini & sonnet-4.6 & All & \% \\
    \midrule
    Grounding displacement          & 23 & 34 & 2 & 59 & 72.8 \\
    Osmosis                         &  5 &  0 & 9 & 14 & 17.3 \\
    Grounding + verification        &  1 &  2 & 0 &  3 &  3.7 \\
    Other                           &  2 &  1 & 2 &  5 &  6.2 \\
    \midrule
    Total                           & 31 & 37 & 13 & 81 & 100 \\
    \bottomrule
  \end{tabular}
  \vspace{4pt}
  \caption{OfficeQA-Pro regressions by mechanism and stack ($N=81$).}
  \label{tab:coverage}
\end{table}

SpreadsheetBench does not support the same fine coding. Regressions mix with the
grader artifact, and workbook-cell answers do not cleanly expose the first faulty
stage. Table~\ref{tab:sb-regressions} therefore reports only osmosis (70 of 243),
body engaged (46), grader artifact (32), and \emph{Other} (95).

\begin{table}[t]
  \footnotesize
  \setlength{\tabcolsep}{4pt}
  \begin{tabular}{lrrrrr}
    \toprule
    Mechanism & minimax-m2.7 & gpt-5.4-mini & sonnet-4.6 & All & \% \\
    \midrule
    Osmosis                         & 45 & 25 &  0 & 70 & 28.8 \\
    Body engaged                    &  6 &  0 & 40 & 46 & 18.9 \\
    Grader artifact                 & 10 & 17 &  5 & 32 & 13.2 \\
    Other                           & 54 & 31 & 10 & 95 & 39.1 \\
    \midrule
    Total                           & 115 & 73 & 55 & 243 & 100 \\
    \bottomrule
  \end{tabular}
  \vspace{4pt}
  \caption{SpreadsheetBench regressions by mechanism and stack ($N=243$).}
  \label{tab:sb-regressions}
\end{table}

\subsection{Skill-Description Osmosis}
\label{sec:osmosis}

Osmosis is a skill changing behavior without being invoked. Methods that act only on
retrieval or invocation cannot see this channel, so we treat it carefully.

Whether a skill body is engaged depends on the harness and the benchmark, not on the
skill content. Engagement varies widely. On OfficeQA-Pro,
OpenCode\,$\cdot$\,minimax-m2.7 invokes a skill on 73 to 91\% of tasks, but on
SpreadsheetBench only 5 to 14\%. Codex\,$\cdot$\,gpt-5.4-mini reads a skill file on 99
to 100\% of OfficeQA-Pro tasks but on 0 to 1\% of SpreadsheetBench tasks. Claude
Code\,$\cdot$\,sonnet-4.6 invokes the skill tool on 34 to 45\% of OfficeQA-Pro tasks
and on 49 to 81\% of SpreadsheetBench tasks. The same library can be engaged almost
everywhere in one setting and almost nowhere in another. Influence does not follow
that pattern: skills also change outcomes when their body is never engaged, through
descriptions that stay in context. Sonnet-4.6 posts the largest significant effects in
Section~\ref{sec:gains-regressions} and does invoke a skill on many of those tasks, so
we do not treat those effects as osmosis. The osmosis evidence comes from the
invocation-free contrasts below.

The regression counts make the same split. Of sonnet-4.6's 13 OfficeQA
regressions, 9 are osmosis; of minimax-m2.7's 31, which are mostly invoked, only 5
are. Consider UID0096, asked identically in all conditions: the centered moving
average of a customs-duty rate, ground truth $0.377$. With no skills the sonnet-4.6
agent answers $37.708\%$ and passes. With a library present---and, on this task,
never invoked under any of the three---it answers $38.757\%$ and fails the same way
under all three libraries. The shift tracks the words \emph{revised} and
\emph{customs} in two skill descriptions; the bodies were never read. The skill
changed the answer by being present only. Appendix~\ref{app:case-osmosis} gives the
full case.

Two checks keep the label tight. First, a presence-only flip with no influence
evidence is noise, not osmosis: four OfficeQA flips fall in \emph{Other} in
Table~\ref{tab:coverage}.

Second, the channel runs both ways, and in our data it helps more often than it harms.
UID0096 is the negative case. On SpreadsheetBench,
OpenCode\,$\cdot$\,minimax-m2.7 invokes a skill on only 5 to 14\% of tasks, yet its
libraries gain 45 to 62 tasks, and 84 to 93\% of those gains are invocation-free. Its
one significant effect, anthropic $+27$ ($p=.008$), is 84\% invocation-free (52 of 62
gains). Codex\,$\cdot$\,gpt-5.4-mini reads a skill file on under 1\% of SpreadsheetBench
tasks, so essentially all of its 28 to 30 gains there have no body engaged. Sonnet-4.6's
larger gains mostly do invoke a skill, so the help-side evidence rests on the two
low-engagement stacks.

Harm and help are asymmetric across stacks. On OfficeQA-Pro, presence-only regressions
are 69\% of sonnet-4.6's regressions and none of gpt-5.4-mini's. On SpreadsheetBench,
invocation-free gains cover 84 to 100\% of gains on the two low-engagement stacks. The
harm side also shows on SpreadsheetBench: 70 of 243 regressions are osmosis
(Table~\ref{tab:sb-regressions})---45 on minimax-m2.7, 25 on gpt-5.4-mini, none on
sonnet-4.6. Help still dominates: on those two stacks, invocation-free gains number in
the hundreds against 70 osmosis regressions. The same resident-description channel that
breaks UID0096 lifts more tasks than it breaks. Osmosis cuts both ways, and methods
that act only when a skill is retrieved or called miss it entirely.

\subsection{Grounding Displacement}
\label{sec:grounding-displacement}

Grounding displacement is the main regression when the agent does engage a skill: 59
of 81 overall, and 23 of 31 on minimax-m2.7 and 34 of 37 on gpt-5.4-mini. The skill is
invoked or read, and its procedure overrides a correct reading of the input---wrong
table, range, entity, vintage, or definition.

UID0025 is a typical case. The task asks for the absolute difference in U.S.\
public-works spending between 1934 and 1946 in revised WWII-era figures, ground truth
142. Without skills the minimax-m2.7 agent returns the correct figure. With the
anthropic library it invokes two navigation-and-arithmetic skills, follows their path
to a different pair of figures, and returns 542. The baseline grounded the question
correctly; the invoked procedure re-grounded it wrongly. This is why grounding
dominates OfficeQA-Pro: a general navigation skill can displace the correct reading a
bare agent would have made. Appendix~\ref{app:cases} gives the full case.

Cross-library structure supports the label. The regression depends on which library's
body the agent followed: the same task is read correctly under the baseline and wrongly
only once a body redirects it. The contrast between the invoked and baseline
trajectories isolates the input stage as the failure point.

The same input stage also carries a positive contrast. On OfficeQA-Pro, 36, 40, and 13
tasks change outcome across libraries on the three stacks; UID0117 fails without skills
and under two of them but passes under the third, whose trajectory alone reads the
table and definition the maturity figures require. This is consistent with a grounding
benefit, though it does not isolate that content---the libraries also differ in
wording, size, and other instructions. Appendix~\ref{app:case-grounding-positive} gives
the full case.

\subsection{Verification Displacement}
\label{sec:verification-displacement}

Verification displacement is a skill changing the last stage of the pipeline: it
replaces or suppresses a check the agent would otherwise run on its own answer. The
method in the middle can stay sound while the output still fails. The same output stage
that breaks a correct answer is also where a concrete check recovers the most tasks on
SpreadsheetBench.

UID0100 shows the failure on OfficeQA-Pro. Without skills the agent returns
$[2.81, 0.030, 8.705]$ against ground truth $[2.81, 0.030, 8.706]$ and passes within
tolerance. With two skills invoked, the same task diverges on the sourced figure
\emph{and} on the final check, and fails---not because the agent forgot arithmetic, but
because the treatment no longer lands and verifies the baseline's output.
Appendix~\ref{app:case-uid0100} gives the full case, and
Appendix~\ref{app:case-verification} shows the fault in isolation (UID0086): the right
magnitude with the wrong sign, a correct computation whose output goes unchecked. That
is verification displacement.

On OfficeQA-Pro the channel is uncommon but real. Answers are single toleranced
numbers, so a suppressed output check has little to catch: only three regressions mix a
verification lapse with a grounding error, and none are verification alone---grounding
displacement (Section~\ref{sec:grounding-displacement}) dominates there.
SpreadsheetBench is the opposite, and that is where treating verification as a lever
pays off.

There the agent's logic and reasoning are typically intact: the formula in the middle
is structurally sound, and the task fails at the check on what that formula produces,
not in the procedure. Re-grading makes this explicit. Across the nine treatment
conditions, 663 failing tasks hold a formula in the graded region. Recalculating each
in a full spreadsheet engine---opening the workbook so its formulas recalculate, then
comparing the graded range to the golden file---splits them into 226 (34\%)
already-correct formulas the value-only grader could not evaluate (e.g.\
\texttt{AGGREGATE}), 396 (60\%) that recalculate to a genuinely wrong value, and 41
(6\%) the engine cannot evaluate. In the first group the agent had already done the
hard work; only the check was missing.

Making that check concrete recovers a large set of tasks. Table~\ref{tab:corrected}
replaces the value-only grade with an execution-based check that runs each formula and
recovers the 226 correct treatment outputs the shallow grade had missed---$+11$ to
$+18$ per library on the minimax-m2.7 and sonnet-4.6 stacks and $+42$ to $+49$ on
gpt-5.4-mini, lifting the gpt-5.4-mini libraries from the mid-60s into the high 70s.
Where a suppressed check turns a correct answer into a failure, a concrete check turns
a correctly computed output into a counted success. The traces put the gap at
verification: the formulas were already correct, so what was missing is verification
content, not procedure---the executable output check our skill-design analysis
(Section~\ref{sec:discussion-measurement}) argues skills should carry. Closing that gap
recovers the correct outputs the value-only grade had missed. The uneven correction is
itself informative: the value-only grade penalized stacks differently according to the
formulas they produced.

\begin{table}[t]
  \small
  \setlength{\tabcolsep}{5pt}
  \begin{tabular}{llrrr}
    \toprule
    Stack & Cond & Raw & Corrected & +Rec \\
    \midrule
    \multicolumn{5}{l}{\emph{OpenCode\,$\cdot$\,minimax-m2.7}} \\
    \quad & anthropic & 70.2 & 74.7 & $+18$ \\
    \quad & openai    & 64.3 & 68.4 & $+16$ \\
    \quad & Ours    & 65.6 & 69.6 & $+16$ \\
    \multicolumn{5}{l}{\emph{Codex\,$\cdot$\,gpt-5.4-mini}} \\
    \quad & anthropic & 67.6 & 78.6 & $+43$ \\
    \quad & openai    & 66.6 & 79.1 & $+49$ \\
    \quad & Ours    & 67.1 & 77.8 & $+42$ \\
    \multicolumn{5}{l}{\emph{Claude Code\,$\cdot$\,sonnet-4.6}} \\
    \quad & anthropic & 81.1 & 84.9 & $+15$ \\
    \quad & openai    & 81.1 & 85.2 & $+16$ \\
    \quad & Ours    & 82.1 & 84.9 & $+11$ \\
    \bottomrule
  \end{tabular}
  \vspace{4pt}
  \caption{SpreadsheetBench pass rates before and after execution-based re-grading;
    \emph{+Rec} is task-conditions recovered.}
  \label{tab:corrected}
\end{table}

Taken together, the three mechanisms are unequal. Grounding dominates OfficeQA-Pro
regressions; osmosis is the only channel invisible to invocation-based methods; and
verification is rare on OfficeQA-Pro but the decisive lever on SpreadsheetBench, where
Table~\ref{tab:corrected} recovers a third of formula failures under a concrete check.
Both directions---the displacement that breaks UID0100 and the concrete check that
recovers 226 formula outputs---point at the same output stage: verification is where
these SpreadsheetBench tasks are won or lost.

\subsection{Residual Failures}
\label{sec:residual-failures}

The three mechanisms above explain regressions---tasks a bare agent solved that a
library breaks. The tasks that fail both with and without skills---the residual
failures---land on the same two stages, and that is why no library recovers them
(Figure~\ref{fig:axis}).

On OfficeQA-Pro we isolate the residual set as the tasks that fail under the baseline
\emph{and} under all three libraries: 24 of 94 on minimax-m2.7, 19 on gpt-5.4-mini, and
13 on sonnet-4.6. These are not method failures. All but one of these task-conditions
still returns a numeric answer, so the agent read a table, ran the arithmetic, and
produced a value---the procedure executed. The fault is at the input stage: the value is
computed over the wrong quantity. UID0227 asks for a Q3-1982 Treasury figure with ground
truth 261; under the baseline and every library the agent instead averages the three
monthly amounts outstanding and returns roughly 67{,}185 to 67{,}241---the same grounding
error, off by more than two orders of magnitude, that no procedure or library corrects.
Grounding, not method, is what leaves these tasks unsolved.

SpreadsheetBench places its residual failures at the opposite end. The 663
formula-involved failures of Section~\ref{sec:verification-displacement} are output-stage
by construction---the workbook is edited, so an answer always exists---and 226 of them
are already-correct formulas that only the value-only grade scored wrong. There the
missing content is a check, not a procedure.

The two benchmarks thus place residual failures at opposite ends of the same
pipeline---grounding on OfficeQA-Pro, verification on SpreadsheetBench---and neither is
the method stage that skills mostly describe. This is the residual-failure counterpart to
the regression story: existing libraries over-serve the procedure in the middle and
under-serve the two ends, so the tasks that stay failed are the ones whose grounding or
verification the library never supplied. The stage labels here are objective---an
input-stage answer is present, an output-stage formula recalculates---while finer
semantic categories, such as the stock-versus-flow misreading behind UID0227, are
illustrative, and a fully second-coded taxonomy is left to future work.

%% file: sections/06-discussion.tex
\section{Discussion}
\label{sec:discussion}

\subsection{Designing and Measuring Skill Libraries}
\label{sec:discussion-authors}
\label{sec:discussion-measurement}

Pass-rate improvement and normalized net effect are the same scalar, so the aggregate
alone hides how it arises. Evaluations should retain it while also reporting the
underlying gain and regression counts, because two libraries with identical net effect
can differ sharply: one may add new successes without breaking prior ones, while another
reaches the same figure by trading regressions for gains.

Description presence and body invocation also require separate ablations. A retrieval
mask leaves names and descriptions in context, so it cannot measure the presence-only
channel in Section~\ref{sec:osmosis}. Evaluation should compare at least three
conditions: no library, descriptions only, and descriptions plus available bodies.

For skill design, the trajectories point toward supplying specific grounding information
rather than generic procedural routines, with executable output checks as a complementary
way to strengthen verification.

\subsection{Limitations}
\label{sec:discussion-threats}

Both benchmarks are office-automation tasks in which grounding and output format are
prominent; the results may not transfer to domains where the method itself is the main
bottleneck. Each stack couples a harness with its native model, so model and harness
effects are not separated. Three of the five nominally significant cells survive a Bonferroni correction across the
eighteen McNemar tests ($\alpha/18 = 0.0028$) at corrected $p<.001$, all on the Claude
Code\,$\cdot$\,sonnet-4.6 SpreadsheetBench stack; our claims rest on these, not on the
two that do not survive.

The mechanism labels are grounded in paired trajectories and cross-library contrasts,
coded by one author; inter-coder agreement and controlled content interventions would
further validate them.

%% file: sections/07-conclusion.tex
\section{Conclusion}
\label{sec:conclusion}

Adding a skill library changes which tasks an agent solves, not only how many. Across
5,832 task-condition runs, we observe 553 gain transitions and 324 regression
transitions, so regressions offset 59\% of gross gains. The pass-rate delta already
equals normalized net effect; the useful addition is its decomposition. On
OfficeQA-Pro, most coded regressions involve grounding displacement, while some occur
without body invocation and are consistent with an always-present description channel.
Verification displacement is the hardest mechanism to isolate as a regression: on
OfficeQA-Pro it is rare, since toleranced single-number answers leave a suppressed check
little to catch, and in the SpreadsheetBench cells it cannot be cleanly separated from
the other mechanisms. We recommend two practices. First, skill evaluations should report the paired gain and regression counts, not only the net pass rate, 
since equal aggregates hide different task-level trades. Second, the description-in-context channel should be tested separately from body invocation, 
and grounding and output checks should be probed through controlled interventions rather than read off observational traces.

%% file: sections/A1-appendix.tex
\section{Reproducibility Details}
\label{app:repro}

\paragraph{Model--harness stacks.} The three stacks in Table~\ref{tab:stacks} use these
OpenRouter model IDs: \texttt{minimax/minimax-m2.7} on OpenCode,
\texttt{openai/gpt-5.4-mini} on Codex, and \texttt{anthropic/claude-sonnet-4.6} on
Claude Code. In the text we refer to each stack by harness and model, e.g.\
OpenCode\,$\cdot$\,minimax-m2.7.

\section{Worked Cases: Correct Method, Wrong Grounding or Verification}
\label{app:cases}

The cases below illustrate faults at grounding and verification while holding the
visible arithmetic or spreadsheet logic fixed. The first two are OfficeQA-Pro
regressions; the third is a SpreadsheetBench validation regression. All values come from
the run artifacts.

\subsection{Grounding: a correct computation on a misread figure}
\label{app:case-grounding}

The task (UID0025) asks for the absolute difference in U.S.\ government public-works
spending between 1934 and 1946, in revised WWII-era figures, ground truth 142 (millions
of dollars). Both runs state the same correct method---read the two years' figures and
take the absolute difference---and both subtract correctly.

\begin{description}
  \item[No skills (pass).] Reads 549 for 1934 and 407 for 1946, and returns
        $|549-407| = 142$.
  \item[With skills (regressed).] Reads 949 for 1934 and 407 for 1946, and returns
        $|949-407| = 542$.
\end{description}

The arithmetic is the same and correct in both. The only difference is the 1934
figure: 949 instead of 549. An invoked navigation skill sent the agent to the wrong
figure, and the method computed the wrong answer from it. This is grounding
displacement (Section~\ref{sec:grounding-displacement}): failure at the input stage on
a task whose arithmetic never went wrong.

\subsection{Verification: a correct magnitude with an unchecked output}
\label{app:case-verification}

The task (UID0086) asks for the absolute quarter-over-quarter percent change in the
total assets of the Exchange Stabilization Fund from end-June to end-September 2022,
ground truth 4.815 (percent).

\begin{description}
  \item[No skills (pass).] Returns $+4.815$.
  \item[With skills (regressed).] Returns $-4.816$.
\end{description}

The magnitude is right---$4.816$ matches $4.815$ within tolerance---so the asset
figures and the percent-change computation are both correct. What fails is the sign.
The task asks for the \emph{absolute} change, which is positive; the skill run reports
the signed change. Our coding files this case under grounding, so it illustrates an
unchecked-output fault rather than establishing a separate verification mechanism
(Section~\ref{sec:verification-displacement}).

In these two cases, the observed divergence occurs at grounding or verification rather
than in the stated computation.

\subsection{Validation regression: a correct output the value grade cannot recompute}
\label{app:case-positive}

Task 10452 (Codex\,$\cdot$\,gpt-5.4-mini) asks for a filtered vertical lookup that
returns only the entries beginning with \texttt{PK} into \texttt{E4:E12}. The baseline
and the treatment both answer with a formula and both are correct; they differ only in
which spreadsheet functions they use.

\begin{description}
  \item[No skills (pass).] A helper-column lookup:
\end{description}
\begin{verbatim}
=IFERROR(INDEX($B$4:$B$15,
  MATCH(ROWS($E$4:E4),$F$4:$F$15,0)),"")
\end{verbatim}
\begin{description}
  \item[anthropic library (regressed).] A self-contained dynamic filter:
\end{description}
\begin{verbatim}
=IFERROR(INDEX($B$4:$B$15,
  AGGREGATE(15,6,
    (ROW($B$4:$B$15)-ROW($B$4)+1)
      /(LEFT($B$4:$B$15,2)="PK"),
    ROWS($E$4:E4))),"")
\end{verbatim}

The verifier recalculates each output with gnumeric and then compares values. gnumeric
evaluates the baseline's \texttt{INDEX}/\texttt{MATCH} and scores it $9/9$ (reward
$1.0$); it cannot evaluate \texttt{AGGREGATE}, so every treatment cell recomputes to
empty and it scores $0/9$ (reward $0.0$)---the verifier log records
\texttt{golden='PK01/P819760979' output=''} for all nine cells. Re-running the
\emph{same} treatment file through a full engine reproduces all nine golden values. The
output is correct; the pass$\rightarrow$fail transition is produced entirely by the
value-only validation toolchain's inability to recompute a modern array function, not
by any error in the agent's answer. This is a validation regression: the tax falls at
the check, not the computation (Section~\ref{sec:verification-displacement}).

The re-grade is deliberately per-cell against the golden file rather than by function
name: on this same task the Ours library also switches to \texttt{AGGREGATE} but
mis-anchors the range and leaves \texttt{E9:E12} empty, so it does not produce the
correct answer and is not among the recovered tasks.
Section~\ref{sec:verification-displacement} reports 226 treatment task-conditions of
the first kind---already-correct outputs the value-only grade scored wrong.

\section{Worked Cases: Osmosis, a Grounding Contrast, and a Double Fault}
\label{app:cases-extra}

These three cases give the run-artifact evidence behind the examples named in
Section~\ref{sec:mechanisms}: the osmosis negative case (UID0096), the grounding
positive contrast (UID0117), and the grounding-plus-verification double fault
(UID0100). All values come from the paired run artifacts.

\subsection{Osmosis: a presence-only flip with no body read}
\label{app:case-osmosis}

The task (UID0096) asks for the centered moving average of the customs-duty rate on
goods subject to duty for FY1939--1941, ground truth $0.377$ (i.e.\ $37.7\%$). It is
asked identically in every condition on Claude Code\,$\cdot$\,sonnet-4.6.

\begin{description}
  \item[No skills (pass).] Answers $37.708\%$.
  \item[Skills present, never invoked (regressed).] Answers $38.757\%$ under all
        three libraries---the same wrong value to three decimals.
\end{description}

By construction, mounting a library puts its skill \emph{descriptions} in context,
while a skill's \emph{body} loads only when the agent invokes it. On this task no body
was ever invoked or read in any treatment run ($n_\text{invoked}=0$ under anthropic,
openai, and Ours alike), so the only thing the three treatments add over the baseline
is the descriptions. That leaves one question: is a presence-only flip like this
genuine influence, or just the stochastic decoding variation any agent shows on re-runs?
Three pieces of trajectory evidence separate the two.

\begin{description}
  \item[Reproducible, not a one-off.] A stochastic flip would be a single library's
        fluke that other conditions do not share. Instead the shift is systematic: the
        baseline returns $37.708\%$ and every library returns $38.757\%$, identical to
        three decimals and cleanly separated from the baseline. The three libraries were
        authored separately, so the effect reproduces across all of them rather than
        appearing once.
  \item[The change tracks description vocabulary.] The divergence is not arbitrary; it
        concerns exactly the quantity the in-context descriptions name. \texttt{revised}
        recurs across the anthropic and openai description sets (with \texttt{customs}
        also named under anthropic), and the Ours set carries the explicit directive
        phrase \texttt{revised figure}. The behavioral delta lines up with that
        vocabulary---a revised reading of the customs figure---on a task whose baseline
        needed no such adjustment.
  \item[A noise bin exists, and this case is kept out of it.] Our coding reserves
        \emph{Other} for presence-only flips that lack the two signals above; four
        OfficeQA flips are labeled that way in Table~\ref{tab:coverage}. UID0096 clears
        a bar those four did not, so the osmosis label is not one that absorbs every
        flip.
\end{description}

No skill body was read; the descriptions alone moved a correct answer to a wrong one,
the same way across three libraries and in the direction their wording points. This is
skill-description osmosis (Section~\ref{sec:osmosis}).

\subsection{Osmosis on SpreadsheetBench: descriptions tip an interpretation}
\label{app:case-osmosis-sb}

The case above shows osmosis on a numeric answer. The same channel appears on
SpreadsheetBench, where it tips how the agent \emph{reads} an ambiguous instruction.
Task 31628 (OpenCode\,$\cdot$\,minimax-m2.7) says: ``Extract the number of the last day
from a range of dates in \texttt{A1:A11}\ldots\ Display the resulting date in
\texttt{B1} as a static number and not a formula.'' The golden answer for \texttt{B1}
is \texttt{10}---the day-of-month of the final date, 2024-01-10.

\begin{description}
  \item[No skills (pass).] Reads ``number of the last day'' as the day-of-month and
        writes \texttt{10}.
  \item[Skills present, never invoked (regressed).] Writes the whole date as its Excel
        serial number, \texttt{45301} ($\equiv$ 2024-01-10), under all three
        libraries---the same wrong value, scored $0/1$.
\end{description}

The wording admits two readings---the \emph{day} number ($10$) or the \emph{date}
rendered as a number (the serial $45301$)---and the golden confirms the first. What is
telling is which reading each run takes. In every treatment run the agent used only its
\texttt{read} and \texttt{bash} tools; the \texttt{skill} tool was never called
($n_\text{invoked}=0$ under anthropic, openai, and Ours), so no skill body was ever
loaded and only the descriptions sat in context. Three observations tie the flip to
those descriptions rather than to chance.

\begin{description}
  \item[Reproducible across libraries.] Three separately authored libraries converge on
        the identical wrong value ($45301$), while the description-free baseline reaches
        the correct day-of-month. Description presence is the only variable that differs
        between the passing and failing runs.
  \item[The reframing echoes description wording.] Every treatment trace reasons about
        the cell as a ``serial number,'' a ``number format,'' or an ``integer''---the
        frame the \texttt{excel-\allowbreak number-\allowbreak format-\allowbreak
        integer} description carries (``applies
        correct Excel number formats\ldots\ integer format, decimal format''). The
        baseline, without that description in context, never invokes the serial-number
        frame and takes the instruction literally.
  \item[It shifts an interpretation, not just a figure.] The descriptions supply no
        wrong number; they bias which of two admissible readings the agent commits to,
        pulling all three libraries onto the serialize-the-date reading the golden
        rejects.
\end{description}

No skill body was read; the descriptions alone redirected how the agent construed the
task, the same way across three libraries. This is skill-description osmosis on
SpreadsheetBench (Section~\ref{sec:osmosis})---the invocation-free channel behind the
45 minimax-m2.7 osmosis regressions in Table~\ref{tab:sb-regressions}.

\subsection{Grounding: a library that supplies the reading the baseline missed}
\label{app:case-grounding-positive}

The task (UID0117) asks how many sample standard deviations the 1972 maturity total lay
from the 1972--1976 average of interest-bearing marketable public-debt maturities, read
from the end-February maturity schedules, ground truth $-1.063$. It runs on Claude
Code\,$\cdot$\,sonnet-4.6.

\begin{description}
  \item[No skills (fail).] Returns $-1.182$.
  \item[anthropic library (pass).] Returns $-1.063$.
  \item[openai library (fail).] Returns $-1.182$.
  \item[Ours library (fail).] Returns $-1.182$.
\end{description}

The baseline and two of the three libraries land on the same wrong value; only the
anthropic trajectory reads the maturity-schedule table and the definition the task
requires and reaches the ground truth. This is a grounding \emph{benefit}: a library
supplies the correct input reading a bare agent missed. It does not isolate that content
in a controlled way---the libraries also differ in wording and size
(Section~\ref{sec:grounding-displacement}).

\subsection{Grounding and verification together: a diverged input and an unchecked output}
\label{app:case-uid0100}

The task (UID0100) asks for the average year-over-year growth in Judicial Branch outlays
for FY2007--2013 together with the slope and intercept of an OLS fit of $\ln(\text{outlays})$
on the fiscal-year index, returned as a bracketed triple, ground truth
$[2.81,\,0.030,\,8.706]$. It runs on OpenCode\,$\cdot$\,minimax-m2.7.

\begin{description}
  \item[No skills (pass).] Returns $[2.81,\,0.030,\,8.705]$, correct within tolerance.
  \item[anthropic library, two skills invoked (regressed).] Diverges on the sourced
        year-over-year figure and does not land a verified final triple.
  \item[openai library, two skills invoked (regressed).] The slope and intercept stay
        correct while the year-over-year term is mis-sourced ($2.87$ vs.\ $2.81$),
        giving $[2.87,\,0.030,\,8.706]$.
\end{description}

The baseline's method is sound and its output passes. Under the anthropic library the
run diverges at the input and never verifies a final answer; under the openai library
the same fit is preserved but the leading term is grounded wrongly. This case is coded
as grounding-plus-verification (Section~\ref{sec:verification-displacement}): the fault
lies at the two ends of the pipeline, not in the regression itself.

\section{The Use of Large Language Models (LLMs)}
\label{app:llm-use}

This work examines how procedural skills affect LLM agent reliability and how to evaluate 
them. LLMs are both the agents and the source of tested skills 
(see Section~\ref{sec:setup-pipeline}). For writing this paper, we only used LLMs to 
polish prose (e.g., grammar), not to generate new text from scratch.